\pdfoutput=1

\documentclass[11pt]{article}

\usepackage[]{acl}
\usepackage{wrapfig,boxedminipage,lipsum}
\usepackage{times}
\usepackage{latexsym}
\usepackage{tcolorbox}
\usepackage{adjustbox}
\usepackage{textcomp}
\usepackage{pgfplots}
\pgfplotsset{width=\linewidth,compat=1.5}
\usepackage[T1]{fontenc}

\usepackage[utf8]{inputenc}

\usepackage{microtype}

\title{Nearest Neighbor Search over Vectorized Lexico-Syntactic Patterns for Relation Extraction from Financial Documents}

\author{Pawan Kumar Rajpoot $^*$ \\
  UtilizeAI Research  \\
  Bangalore, India   \\
  \texttt{pawan.rajpoot2411@gmail.com} \\\And
  Ankur Parikh $^*$ \\
  UtilizeAI Research \\
  Bangalore, India\\
  \texttt{ankur.parikh85@gmail.com} \\}

\begin{document}
\maketitle
\begin{abstract}
Relation extraction (RE) has achieved remarkable progress with the help of pre-trained language models. However, existing RE models are usually incapable of handling two situations: implicit expressions and long-tail relation classes, caused by language complexity and data sparsity. Further, these approaches and models are largely inaccessible to users who don’t have direct access to large language models (LLMs) and/or infrastructure for supervised training or fine-tuning. Rule-based systems also struggle with implicit expressions. Apart from this, Real world financial documents such as various 10-X reports (including 10-K, 10-Q, etc.) of publicly traded companies pose another challenge to rule-based systems in terms of longer and complex sentences. In this paper, we introduce a simple approach that consults training relations at test time through a nearest-neighbor search over dense vectors of lexico-syntactic patterns and provides a simple yet effective means to tackle the above issues. We evaluate our approach on REFinD and show that our method achieves state-of-the-art performance. We further show that it can provide a good start for human in the loop setup when a small number of annotations are available and it is also beneficial when domain experts can provide high quality patterns. Our code is available at \footnote{https://github.com/pawan2411/PAN-DL\_Refind}.
\end{abstract}
\def\thefootnote{*}\footnotetext{Equal Contribution}\def\thefootnote{\arabic{footnote}}
\section{Introduction}

Relation extraction (RE) from text is a fundamental problem in NLP and information retrieval, which facilitates various tasks like knowledge graph construction, question answering and semantic search. Recent studies \cite{zhang-etal-2020-minimize, zeng2020copymtl, lin-etal-2020-joint, wang-lu-2020-two, cheng-etal-2020-dynamically, zhong-chen-2021-frustratingly}  in supervised RE take advantage of pre-trained language models (PLMs) and achieve SOTA performances by fine-tuning PLMs with a relation classifier. However, \cite{wan-etal-2022-rescue} observes that existing RE models are usually incapable of handling two RE-specific situations: implicit expressions and long-tail relation types. 

Implicit expression refers to the situation whereas relation is expressed as the underlying message that is not explicitly stated or shown. 

In Figure 1, relation "acquired\_by(organization, organization)" occurs implicitly. Such underlying messages can easily confuse the relation classifier. 
\begin{figure}
\includegraphics[width=\linewidth,scale=5,]{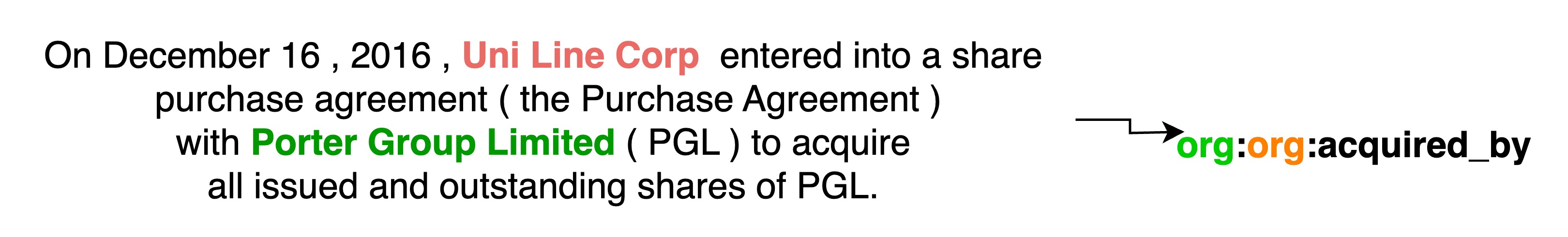}
\caption{Relation Extraction example, here both organizations are connected with "acquired by" relation.}
\label{fig:Entity and Relation types in REFinD dataset}
\end{figure}

The other problem of long-tail relation classes is caused by data sparsity in training. For example, the REFinD dataset \cite{kaur2023refind} comprises 45.5 \% of the no\_relation instances. The most frequent class in the dataset - “per:title:title” has 4,468 training examples, while over 14 out of 22 classes have less than 500 examples. The majority class can easily dominate model predictions and lead to low performance on long-tail classes. 


Recently, ICL (In-Context Learning) based approach \cite{wan2023gptre} is utilized for RE tasks. The approach achieves improvements over not only existing GPT-3 baselines, but also on fully-supervised baselines even with only a limited number of demonstrations provided in the prompt. Specifically, it achieves SOTA performances on the Semeval and SciERC datasets, and competitive performances on the TACRED and \cite{zhang2017tacred} ACE05 datasets. \cite{rajpoot2023gptfinre} utilized the GPT-4 under ICL framework on REFinD and achieved 3rd rank in the shared task. 

However, retrieval of examples to demonstrate is a key factor in the overall performance on these pipelines. Finding efficient demonstrates often relies on learning-based retrieval \cite{ye2023compositional, rubin-etal-2022-learning}. These learning-based retrievers use annotated data and a LLM. This type of retrieval strategy comes with the increased cost (API, infrastructure etc.), time as more experiments are required because most LLMs are black box and it also needs special expertise. 

Apart from the implicit expression challenge mentioned above, REFinD poses another challenge to rule-based systems in terms of longer and complex sentences. For example, \cite{kaur2023refind} cites that the average sentence length in the REFinD dataset is 53.7 while the average sentence length in the TACRED dataset \cite{zhang-etal-2017-position} is 36.2. Further, As per \cite{kaur2023refind}, REFinD includes more complex sentences than TACRED, with an average entity-pair distance of 11, compared to 8 in TACRED. Because of this, writing rules at surface text level is a challenge. Hence, rules at lexico-syntactic level is the need of the hour. However, strict matching of these rules can yield high precision but low recall result due to accuracy of syntactic parsing. Hence, a robust fuzzy pattern matching system is required.      

Inspired by recent studies \cite{wan-etal-2022-rescue, khandelwal2019generalization, guu2020retrieval, meng2021gnn} using k-Nearest Neighbor to retrieve diverse expressions for language generation tasks, we introduce a simple but effective approach that consults training relations at test time through a nearest-neighbor search over dense vectors of lexico-syntactic patterns and provides a simple yet effective means to tackle the above issues. Our method achieves an improvement of 1.18\% over baseline (F1-score - 0.7516). We achieved our results using commodity hardware within a day. That’s why our approach is easier to deploy, lightweight and fast. We further show that our approach can provide a good start (F1-score of 0.5122) for human in the loop setup when a small number of annotations (approx. 10\% of training data) are available and it is also beneficial (F1-score of 0.6939 with approx. 10\% of training data) when domain experts can provide high quality patterns.  

\section{Preliminary Background}
\subsection{Task Definition}
Let C denote the input context and e1 in C, e2 in C denote the pair of entity pairs. Given a set of predefined relations classes R, relation extraction aims to predict the relation y in R between the pair of entities (e1, e2) within the context C,or if there is no predefined relation between them, predict y="no relation".
\subsection{Data}
The REFinD dataset \cite{kaur2023refind} is the largest relation extraction dataset for financial documents to date. Overall REFinD contains around 29K instances and 22 relations among 8 types of entity pairs. REFinD is created using raw text from various 10-X reports (including 10-K, 10-Q, etc.broadly known as 10-X) of publicly traded companies obtained from US Securities and Exchange Commission.

\section{Nearest Neighbor Search over Vectorized Lexico-Syntactic Patterns}
\subsection{Generating Lexico-Syntactic Patterns}
We replaced words representing entities of interest with their entity types given in the dataset. 

Instead of conducting nearest neighbor search on a complete sentence, we applied Spacy Dependency Parser\footnote{https://spacy.io/} and considered the shortest dependency path (henceforth SDP) between two entities to deal with long and complex sentences with the intuition that considering all sentence words can do more harm in search. SDP is essential for relationship identification in most cases. 

We apply Spacy NER on REFinD sentences and replace actual named entities with their types to create higher-level patterns. 

We also enriched all SDP words with their Dependency Labels to utilize structure information in our search.

For each sentence, we create 4 patterns: 1. SDP words only (SDP) 2. SDP words with named entities replaced with their types (SDP-NER) 3. SDP words enriched with their Dependency Labels (SDP-DEP) 4. SDP words with named entities replaced with their types and also enriched with their Dependency Labels (SDP-DEP-NER). Example patterns are shown in Figure 2. 
\begin{figure}
\includegraphics[width=\linewidth,scale=5,]{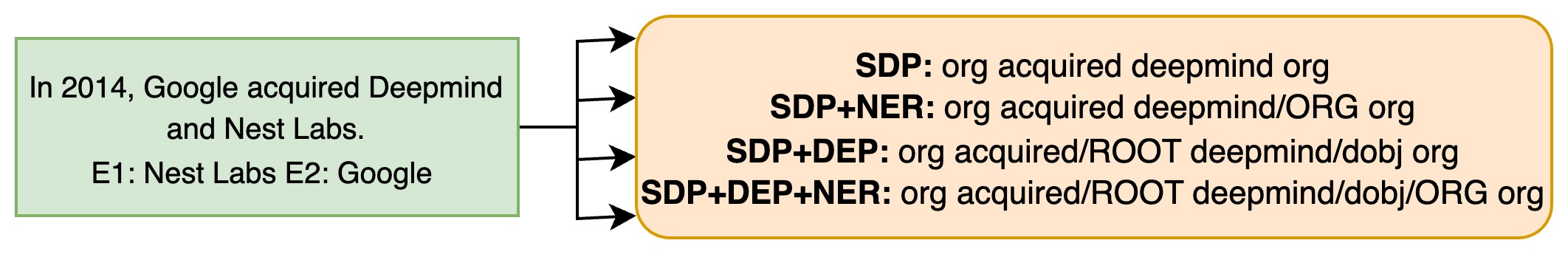}
\caption{Patterns extracted by our pipeline}
\label{fig:Patterns}
\end{figure}
\subsection{Generating Dense Vectors for Lexico-Syntactic Patterns}
We converted all 4 types of Lexico-Syntactic Patterns into Dense Vectors as it performs better than Sparse Vectors. To create a vector, we employed an all-mpnet-base-v2\footnote{https://huggingface.co/sentence-transformers/all-mpnet-base-v2} sentence encoder. We also created vectors for original sentences using the encoder.

\subsection{Creating Class Specific Indices}
For each pattern type mentioned above, we created 21 dense vector indices each representing a relationship class except 'no\_relation' class. We split 'no\_relation' training data instances into 8 splits as per entity-type pairs such as "Person-Organization", "Organization-Organization" etc. and created indices for each split. In this way, there are 29 indices in total for each pattern type. Each element of the index represents a vectorized lexico-syntactic pattern for each training example. For around 11.89\% of the training sentences, we faced issues in generating dependency tree and/or SDP. To deal with this, we also created another 29 indices containing dense vectors for original sentences.

\subsection{Conducting Nearest Neighbor Search}
After configuring lexico-syntactic pattern type and value of K, Given a test sentence and an entity-type pair, we first create a vector representing its lexico-syntactic pattern obtained using steps described above. With the entity-type pair, appropriate relation class indices are selected for search. The pattern vector is searched in every appropriate class index using cosine similarity and top K vectors from each class index are obtained. The similarity scores of each of these top K vectors are averaged and the class having the highest similarity score is selected. In the case of syntactic parsing failures, as a fallback strategy, a vector of the original sentence is created and is used against class specific sentence indices in search the same way as mentioned above.

\begin{table}
\centering
\begin{adjustbox}{width=\columnwidth,center}
\begin{tabular}{lcr}
\hline
\textbf{Pattern} & \textbf{K} & \textbf{F1-score}\\
\hline
{SDP} & {14}  & {0.7552} \\
{SDP-NER} & {12}  & {0.7538} \\
{SDP-DEP} & {11}  & {0.7610} \\
{SDP-DEP-NER} & {14}  & {\textbf{0.7634}} \\
\hline
{Winner on leaderboard (baseline) \footref{fnlabel}}&{-} & {0.7516 }\\
\hline
\end{tabular}
\end{adjustbox}
\caption{Comparison on performance on REFinD dev data}
\label{tab:train_metrics}
\end{table}
\begin{center}
\begin{tikzpicture}
\begin{axis}[
    title={Figure 3: Sensitivity Analysis},
    ylabel={F1-Score},
    xlabel={K},
    xmin=0, xmax=20,
    ymin=0.5, ymax=1,
    xtick={0,2,4,6,8,10,12,14,16,18,20},
    ytick={.60,.80,1},
    legend pos=north west,
    ymajorgrids=true,
    grid style=dashed,
]

\addplot[
    color=blue,
    mark=square,
    ]
    coordinates {
    (1,0.6867301882407623)(2,0.7194980246339763)(3,0.7348361608180339)(4,0.7434348129212177)(5,0.7464559609574716)(6,0.7504066930048803)(7,0.7527306530327679)(8,0.7538926330467116)(9,0.7564489890773879)(10,0.7571461770857542)(11,0.7606321171275854)(12,0.7606321171275854)(13,0.7622588891471067)(14,0.7634208691610506)(15,0.762956077155473)(16,0.767184731582617)(17,0.7617940971415292)(18,0.762956077155473)(19,0.7622588891471067)(20,0.7613293051359517)
    };
    \legend{SDP+DEP+NER}
    \label{pgfplots:plot1}
\node (mark) [draw, red, circle, minimum size = 2pt, inner sep=5pt, thick] 
      at (axis cs: 14,0.7634208691610506) {};
\end{axis}
\end{tikzpicture}
\end{center}

\section{Experiment Settings}
\subsection{Dataset}
The REFinD dataset \cite{kaur2023refind} released with the shared task is a part of "Knowledge Discovery from Unstructured Data in Financial Services" (KDF) workshop which is collocated with SIGIR 2023. There are 20070, 4306 and 4300 instances of training data, development data and public test data respectively. The organizers have released training data, development data and public test data with gold labels but haven't released private test data with gold labels. Because of that, we are not able to benchmark our system against the winners of the shared task. Since, leaderboard \footnote{https://codalab.lisn.upsaclay.fr/competitions/11770\label{fnlabel}} and gold labels on development data is available, we have benchmarked our approach against the leaders of development data. We have used training data and public test data to create class specific indices to perform nearest neighbor search for development data sentences.
\subsection{Hardware Resources}
We have used a laptop with 16GB RAM and Intel® Core™ i7-7500U CPU @ 2.70GHz × 4 CPU to produce these results.
\subsection{Efforts}
Given the dataset, all setup and experiments are conducted within a day.



\section{Results}
We conducted experiments with 4 different pattern types. To find the best value of K, we have created a 10\% split from the training data and experimented with different values of K (1 to 20). During evaluation, we faced issues in generating dependency tree and/or SDP for around 8.7\% instances and for those instances, indices containing sentence vectors were used as fallback strategy. The results in Table 1 show that our best F1-score is 0.7634 for SDP-DEP-NER pattern and K=14 (Top K vectors) and our method shows improvement of 1.18\% over baseline. Figure 3 shows how sensitive this approach is with respect to different values of K. This confirms our intuition that there is value in utilizing vectorized lexico-syntactic patterns to deal with long and complex sentences and implicit expressions. Further, splitting instances as per the class and performing lazy classification over these splits can help in dealing with the dataset with long-tail relation classes. 

To explore the effectiveness of our approach in human in the loop situation, we conducted a few experiments as shown in Figure 4. We randomly selected N patterns per class from the training data and built indices with those patterns only. We tried different values of N. With N=100 and K=1 (derived from dev split), we achieved an F1-score of 0.5122 with around 10\% of the original training data. It shows that the vectorized lexico-syntactic patterns and the cosine similarity based scoring can be a good start to label similar instances when the number of annotations are less. This method can be used in human in the loop setup to either filter out similar instances (explore) or to find similar instances (exploit) for further human review/annotation.

To explore the effectiveness of our approach when domain experts are available and can provide high quality patterns specially for the task like this which is restricted to a particular domain, types of documents, types of entities and a handful of relations, we conducted a few experiments as shown in Figure 4. To approximate this experiment, we selected N training patterns from each class which occurs frequently in Top K search when development patterns are classified correctly. We call these patterns the Most-Frequent Patterns. We built indices with those patterns only. As shown in Figure 4, with N=100 and K=4 (derived from dev split), we achieved an F1-score of 0.6939 (6.95\% less than the best result) with around 10\% of the original training data. It shows that it can bridge gaps quickly with a small amount of high quality patterns.       
\begin{center}
\begin{tikzpicture}
\begin{axis}[
    title={Figure 4: Training Patterns Selection},
    ylabel={F1-Score},
    xlabel={Number of Patterns per class},
    xmin=0, xmax=100,
    ymin=0, ymax=1,
    xtick={0,25,50,75,100},
    ytick={.30,.50,.70,1},
    legend pos=north west,
    ymajorgrids=true,
    grid style=dashed,
]

\addplot[
    color=blue,
    mark=square,
    ]
    coordinates {
        (100,0.5122007901464095)(75,0.5054613060655356)(50,0.484545665814548)(25,0.453636997443644)
    };
    \legend{Randomly selected Patterns}

\addplot[
    color=green,
    mark=square,
    ]
    coordinates {
        (100,0.6939344643272136)(75,0.6739484080873809)(50,0.6544271438531257)(25,0.5619335347432024)
    };
    \legend{Randomly selected; K=1, Most-Frequent; K=4}
    \label{pgfplots:plot2}
\end{axis}
\end{tikzpicture}
\end{center}
\section{Conclusion}
Our approach consults training relations at test time through a nearest-neighbor search over dense vectors of lexico-syntactic patterns. We evaluated our approach on REFinD and show that our method achieves state-of-the-art performance without any direct access to large language models (LLMs) or supervised training or fine-tuning or any handcrafted rules. We achieved our results using a commodity hardware within a day. That’s why our approach is easier to deploy, lightweight and fast. We further explores that our approach can provide a good start for human in the loop setup when a small number of annotations are available and it can be also beneficial when domain experts can provide high quality patterns.

\section{Limitations}
Since our method is based on nearest neighbors search, it's sensitive to the value of K. Furthermore, our method is also very sensitive to syntactic parsing and NER. Our vectors are not optimal representations because our syntactic patterns are not a natural fit for the sentence encoder.  

\bibliography{anthology,custom}
\bibliographystyle{acl_natbib}

\appendix



\end{document}